\documentclass{article}

\PassOptionsToPackage{numbers}{natbib}
\usepackage{natbib}
\usepackage[preprint]{neurips_2023}

\usepackage{microtype}
\usepackage{graphicx}
\usepackage{subfigure}
\usepackage{caption}

\graphicspath{{images/}}
\usepackage{dblfloatfix}
\usepackage{pdfpages}
\usepackage{algorithmic}

\usepackage{amsmath}
\usepackage{amssymb}
\usepackage{mathtools}
\usepackage{amsthm}

\usepackage{enumitem}
\setitemize{noitemsep,topsep=0.1pt,parsep=0.1pt,partopsep=0.1pt}

\usepackage{pdfpages}

\usepackage{setspace}
\usepackage{etoolbox}

\BeforeBeginEnvironment{equation}{\begin{singlespace}\setstretch{0.5}}
\AfterEndEnvironment{equation}{\end{singlespace}\noindent\ignorespaces}
\BeforeBeginEnvironment{align}{\begin{singlespace}\setstretch{0.5}}
\AfterEndEnvironment{align}{\end{singlespace}\noindent\ignorespaces}
\BeforeBeginEnvironment{gather}{\begin{singlespace}\setstretch{0.2}}
\AfterEndEnvironment{gather}{\end{singlespace}\noindent\ignorespaces}

\theoremstyle{plain}

\theoremstyle{definition}

\theoremstyle{remark}

\usepackage[textsize=tiny]{todonotes}
\usepackage{tikz}
\usetikzlibrary{shapes,arrows,calc,decorations.markings,positioning,3d}

\usepackage{xcolor}    
\PassOptionsToPackage{gray}{xcolor}
\definecolor{Gray}{gray}{0.6}
\definecolor{LightGray}{gray}{0.7}
\definecolor{LighterGray}{gray}{0.8}
\usepackage{colortbl}

\usepackage[utf8]{inputenc} 
\usepackage[T1]{fontenc}    
\usepackage[hyperfootnotes=false]{hyperref}       
\usepackage[capitalize,noabbrev]{cleveref}
\crefname{type}{singular}{plural}
\crefname{figure}{Fig.}{Figs.}
\usepackage{url}            
\usepackage{booktabs}       
\usepackage{amsfonts}       
\usepackage{nicefrac}       
\usepackage{microtype}      

\usepackage[para]{footmisc}

\title{Towards Better Orthogonality Regularization with Disentangled Norm in Training Deep CNNs}

\begin{document}

\author{
    Changhao Wu\footnotemark[1]\footnotemark[2]
    \And 
    Shenan Zhang\footnotemark[2]
    \And 
    Fangsong Long\footnotemark[2]
    \And 
    Ziliang Yin\footnotemark[2]
    \And 
    Tuo Leng\footnotemark[3]
}

\maketitle

\footnotetext[1]{Main Contribution}
\footnotetext[2]{Independent Researcher}
\footnotetext[3]{Shanghai University}

\begin{abstract}

Orthogonality regularization has been developed to prevent deep CNNs from training instability and feature redundancy. 
Among existing proposals, kernel orthogonality regularization enforces orthogonality by minimizing the residual between the Gram matrix formed by convolutional filters and the orthogonality matrix.

We propose a novel measure for achieving better orthogonality among filters, which disentangles diagonal and correlation information from the residual. 
The model equipped with the measure under the principle of imposing strict orthogonality between filters surpasses previous regularization methods in near-orthogonality. 
Moreover, we observe the benefits of improved strict filter orthogonality in relatively shallow models, but as model depth increases, the performance gains in models employing strict kernel orthogonality decrease sharply.

Furthermore, based on the observation of the potential conflict between strict kernel orthogonality and growing model capacity, 
we propose a relaxation theory on kernel orthogonality regularization. The relaxed kernel orthogonality achieves enhanced performance on models with increased capacity, 
shedding light on the burden of strict kernel orthogonality on deep model performance.

We conduct extensive experiments with our kernel orthogonality regularization toolkit on ResNet and WideResNet in CIFAR-10 and CIFAR-100. 
We observe state-of-the-art gains in model performance from the toolkit, which includes both strict orthogonality and relaxed orthogonality regularization, 
and obtain more robust models with expressive features. These experiments demonstrate the efficacy of our toolkit and subtly provide insights into the often overlooked challenges posed by strict orthogonality, 
addressing the burden of strict orthogonality on capacity-rich models.

\end{abstract}

\section{Introduction}\label{submission}

Despite the significant success of deep convolutional neural networks \citep{ImagenetDCNN, VGG,resnet,efficientNet, LKCNN}, 
the problems of vanishing gradient \citep{longtermvanishing,vanishinggradients}, feature statistic shifts \citep{BN}, 
and overgrowth saddle points \citep{CNNsaddlepoint} still shadow the training of deep convolutional neural networks. 
To alleviate these training problems, various techniques have been proposed: parameter initialization \citep{solution1}, 
normalization of internal activations \citep{BN}, residual learning \citep{resnet, solution2, residual1}, and orthogonality regularization.
Kernel orthogonality regularization, one approach in orthogonality regularization, is designed by enforcing the gram matrix to be orthogonal \citep{hard-f-orth-2,hard-f-orth_1}. 

In contrast to the prevailing focus on measuring the distance to strict orthogonality \citep{orth-init-and-orth-f-norm,srip} 
and the influence of the isometry property on training \citep{isonet, stiefel_manifold_orth}, 
our work unveils a less conspicuous yet crucial barrier to achieving better orthogonality regularization. 
We posit that the need to prioritize minimizing task loss can render the optimization objective for strict orthogonality regularization intractable, 
which incurs a gap that leads to traditional measures performing less effectively.

\begin{figure}[h]
    \centering
    \includegraphics[width= 0.5\textwidth]{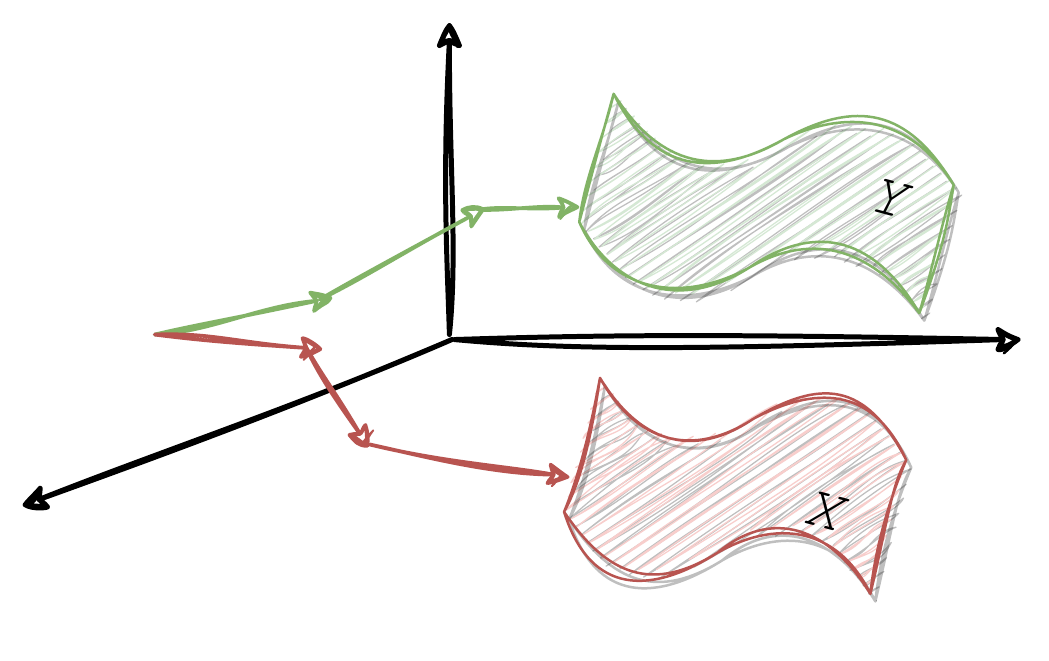}
    \label{inaccessible_orth}
    \caption{\textbf{Inaccessible Orthogonality: } Ideally, the typical SGD trajectory of orthogonality optimization should follow the \textbf{\emph{red}} trace towards $X$, 
    where filters are mutually orthogonal. However, in the context of orthogonality regularization, the task loss dominates the optimization, 
    guiding the \textbf{\emph{green}} SGD trajectory towards $Y$ and ensuring minimal task loss. 
    In \cref{near-orth-table}, we show the greater the model's capacity, the less likely it is for $X$ and $Y$ to overlap. 
    }
\end{figure}

To address this problem, we develop disentangled orthogonality regularization, which, as demonstrated by our extensive experiments, 
outperforms existing orthogonality regularizations in terms of near-orthogonality. In terms of model performance, 
relatively shallow models equipped with disentangled orthogonality demonstrate appreciable gains. 
However, the performance gain diminishes significantly with increasing network depth. \citet{isonet} reported a similar anomaly, 
raising a critical question: should we adhere to strict orthogonality regularization? 

In response to this issue, our disentangled norm serves as a valuable lens to investigate the relationship between near-orthogonality and model performance gains. 
While previous design principles suggest that better near-orthogonality leads to better performance, our strict disentangled orthogonality achieves the best near-orthogonality among existing regularization.
However, we observe that performance gains diminish in deeper models, despite the improved near-orthogonality. This highlights the need to rethink the role of strict orthogonality in deep networks and explore alternative approaches.
Based on this insight, we propose a relaxation theory and, in accordance with this theory, develop a relaxation variant of the disentangled orthogonality regularization.

Remarkably, this relaxation variant of our disentangled orthogonality regularization not only attains state-of-the-art performance but also reveals an intriguing phenomenon. 
By opting to impose strict orthogonality on the transition dimension, which is theoretically efficient for data representation, 
the relaxed variant underscores the advantages of deviating from the previous principle of strict orthogonality regularization. 
In doing so, it effectively compensates for the suboptimal performance of strict orthogonality on the background space in deeper networks.


\section{Related works}

The benefits of orthogonality filters were first researched in recurrent neural networks (RNNs) to alleviate gradient vanishing or exploding problems \cite{rnn-orth1,rnn-orth2,rnn-orth3}.
To utilize cheap computation, \cite{cheap_orth} proposed parameterization from exponential maps. The comparison of soft and hard orthogonality in RNNs is discussed in \cite{hard-soft-orth}.
The advantages of stabilizing the training of CNNs are studied in \cite{cnn-orth1,srip,orth-init-and-orth-f-norm}. How to maintain the orthogonality property in training CNNs is investigated in \cite{hard-f-orth_1,hard-f-orth-2}
using Stiefel manifold-based optimization methods. Orthogonality regularization is reported to improve the training of image generation\cite{gan1,gan2,GAN_spec}.

Imposing semi-orthogonality on the transformation matrix of the network shows favorable outcomes in experiments \cite{hard-f-orth_1,hard-f-orth-2,orth-init-and-orth-f-norm,isonet}. 
Previous pieces of literature explore how to measure the residual between the Gram matrix and identity matrix precisely \cite{srip, stiefel_manifold_orth, orth-init-and-orth-f-norm}, 
and some researchers study how to make approximations in other well-propertied spaces \cite{ocnn}.

\section{\textbf{Disentangled orthogonality and relaxation theory}}

\subsection{\textbf{Preliminary}}

In this section, we first review existing kernel orthogonality regularizations to provide a solid foundation for our discussion. 
Next, we introduce the disentangled norm, offering a more effective approach to enforcing strict kernel regularization in optimization. 
Following the discussion on strict disentangled orthogonality, we emphasize the necessity of developing a relaxation theory on kernel matrix. 
This insight leads us to propose a relaxed version of disentangled orthogonality regularization, 
which addresses the limitations of its strict counterpart in certain scenarios.

We shall first establish a unified notation and clarify the terminology used in the context of kernel orthogonality regularization:

\begin{itemize}[leftmargin=10pt, itemsep=4pt]
\item Convolution filters or kernel matrix $K$: The transformation matrix of convolution layers, denoted as $K$, has dimensions $R^{o \times i \times k_h \times k_w}$, where the abbreviations represent the number of output channels, input channels, height of the convolution kernel, and width of the convolution kernel, respectively.
The kernel matrix $K$ can be reshaped to $R^{o \times (i \times k_h \times k_w)}$ as follows:

\begin{equation}
K = \left(\begin{array}{ccc}
  \rule[0.5ex]{3.0ex}{0.5pt} & k_1 & \rule[0.5ex]{3.0ex}{0.5pt}  \\
    & \vdots &  \\
    \rule[0.5ex]{3.0ex}{0.5pt}  & k_o & \rule[0.5ex]{3.0ex}{0.5pt}
  \end{array}\right),  \text{where } k_i \text{ induce a linear map } k_i^{\top}:\left\langle k_i, \cdot\right\rangle \mapsto R
\end{equation}

This transformation maps the stacked input patches(so-called background space in the following context) to the output channel space $R^{o}$.
\item Gram matrix of the kernel matrix: The Gram matrix is denoted as $\operatorname{Gram}_{o \times o}=K K^{\top}$. The orthogonality of the kernel matrix specifically refers to $K K^{\top} = I_{o \times o}$. 
Strict orthogonality regularization is defined as enforcing the whole gram matrix approaching orthogonality: 

\begin{equation}
  K K^{\top} \longrightarrow I_{o \times o}
\end{equation}

with better strict orthogonality implying that the measure of $K K^{\top} - I_{o \times o}$ is smaller.
\item Over-determined/Less-determined: These terms describe the relationship between the rows and columns in the reshaped kernel matrix. A kernel with $o \geq (i \times k_h \times k_w)$ is defined as less-determined. 
Strict orthogonality is theoretically inaccessible to an over-determined kernel matrix.

\end{itemize}

\subsection{\textbf{Strict kernel orthogonality regularization}}

\subsubsection{Frobenius norm: "Entry-wise" matrix norms orthogonality regularization}

In this part, we introduce the Frobenius norm orthogonality regularization in previous works. 
Frobenius orthogonality regularization aims to optimize the Gram matrix by minimizing the Frobenius norm between the orthogonality and Gram matrix, 
driving it towards zero \cite{orth-init-and-orth-f-norm, stiefel_manifold_orth}:

\begin{gather}
    \left\|K K^{\top}-I_{o \times o}\right\|_F \rightarrow 0        
\end{gather}

\citet{kaist_orth_survery} introduced an improvement to the Frobenius orthogonality regularization by replacing the squared average to balance the loss of different hierarchy structures. 
This improved version of Frobenius orthogonality regularization provides a more balanced approach for handling layers with different filter numbers, 
 $ |\mathcal{W}| = \sqrt{o_i}$, where $o_i \in \text{set} \{o_1, o_2,...  \}$ represents the set of respective out-channels in different blocks. 
In this context, we adopt the form of squared mean of the distance of $K K^{\top}-I_{o \times o}$:

\begin{gather}
  \frac{\left\|K K^{\top}-I_{o \times o}\right\|_F}{|\mathcal{W}|} \rightarrow 0        
\end{gather}

\subsubsection{SRIP: Vector 2-norm orthogonality regularization}

In this part, we introduce the Spectral Restricted Isometry Property Regularization (SRIP) \cite{srip}, 
a method that replaces the Frobenius norm orthogonality regularization with the spectral norm. 
This change significantly enhances the network's generalization ability \cite{spec_norm, GAN_spec, wiki:Matrix_norm} and has led to state-of-the-art performance upon its publication.

Instead of approximating the zero matrix with $K$, SRIP enforces the residual between the Gram matrix and the identity matrix, 
$K K^{\top} - I$, to approximate the zeros matrix under the spectral norm:

\begin{gather}
  \|K K^{\top}-I\|_2=\sup _{x \neq 0} \frac{\|(K K^{\top}-I) x\|_2}{\|x\|_2} =\sigma_{\max }(K K^{\top}-I) \rightarrow 0        
\end{gather}

Due to the high computational complexity of the actual largest eigenvalue, SRIP approximates it using a two-step power iteration:

\begin{gather}
u \leftarrow\left(K K^{\top}-I\right) v, v \leftarrow\left(K K^{\top}-I\right) u, \sigma\left(K K^{\top}-I\right) \leftarrow \frac{\|v\|}{\|u\|} .
\end{gather}

A crucial yet often overlooked factor for the success of the spectral norm is achieving a more balanced ratio between the diagonal norm and the correlation triangle.

\subsubsection{Disentangled norm on strict orthogonality}

In this section, our goal is to derive the disentangled norm.
However, before doing so, we discuss the motivation for applying strict orthogonality regularization. 
We observe that the strict orthogonality regularization pushes the Gram matrix $K K^{\top}$ of the kernel matrix towards strict orthogonality, 
which has three primary effects:

\begin{itemize}[leftmargin=10pt, itemsep=4pt]
\item \textit{Correlation:} Strict orthogonality enforces zero correlation in the Gram matrix, 
indicating no correlation among the filters. This property allows the kernel matrix to effectively avoid filter redundancy or rank collapse in its linear span.
\item \textit{Diagonal:} Strict orthogonality imposes an all-one diagonal in the Gram matrix. 
In combination with the zero correlation. This condition implies that all filters can map isometrically. 
In practice, however, achieving isometric kernel matrix is hard \cite{isonet}. 
\item \textit{Fair Mapping:} During optimization, the variance of the filter linear map is constrained. 
Maintaining a diagonal with low length variance helps prevent suboptimal filter usage, 
where a filter with a significantly smaller norm than the normal level in a less-overdetermined convolutional layer might have a weak output feature, 
despite being perpendicular to the other filters.
\end{itemize}
  
With these insights in mind, we proceed to derive the disentangled norm, which offers a more effective approach to strict orthogonality during optimization. 
We now disentangle the diagonal and correlation information from the Gram matrix $K K^{\top}$. Since the Gram matrix $K K^{\top}$ is a real symmetric matrix, 
it suffices to consider the lower triangular triangle and the diagonal:

\begin{align}
\operatorname{LowerTriangular}(K K^{\top})= \left[\begin{array}{cccc}
  0 & 0 & \cdots & 0 \\
  k_2^{\top}k_1 & 0 & & 0 \\
  \vdots & & \ddots & \vdots \\
  k_n^{\top}k_1 & k_n^{\top}k_2 & \cdots & 0
  \end{array}\right]+\left[\begin{array}{ccc}
  k_1^{\top}k_1  & \cdots & 0 \\
  \vdots         & \ddots & \vdots \\
  0              & \cdots & k_n^{\top}k_n
  \end{array}\right]
\end{align}

Since the correlation between two filters $\operatorname{Corr}\left(k_i, k_j\right)=\frac{\left\langle k_i, k_j\right\rangle}{\|k_i\| \|k_j\|}$ 
can directly measure the orthogonality extent between them regardless of the influence from their norms, 
we believe it is more appropriate to apply the lower triangle of the correlation matrix to compute the correlation loss: 

\begin{gather}
  \left\|\left[\begin{array}{cccc}
    0 & 0 & \cdots & 0 \\
    \operatorname{Corr}\left(k_2, k_1\right) & 0 & \ddots & 0 \\
    \vdots & & \ddots & \vdots \\
    \operatorname{Corr}\left(k_n, k_1\right) & \operatorname{Corr}\left(k_{n}, k_2\right) & \cdots & 0
    \end{array}\right]- \textbf{0}_{o \times o}\right\| _F 
    +
    \lambda \left\|\left[\begin{array}{c}
    k_1^{\top} k_1 \\
    \vdots \\
    k_n^{\top} k_n
    \end{array}\right]-\textbf{1}_{o \times 1}\right\|_F \rightarrow 0  
\end{gather}

$\lambda$ is a balance coefficent between correlation loss and diagonal loss. On the computation complexity, 
we first derive the norm of filters from the kernel matrix and subsequently normalize the filters. 
This allows us to obtain the lower triangle of the correlation matrix directly from the Gram matrix of normalized filters. 
By focusing on the lower triangle of the Gram matrix, our method effectively reduces computation by half compared to the original Frobenius orthogonality approach.

\subsection{Relaxation theory, relaxed disentangled orthogonality}

\subsubsection{Relaxation on over-determined layers}

In this section, we propose the relaxation theory on kernel orthogonality regularization.
Starting with the previously defined over-determined convolutional layers, these layers are characterized by having a greater number of filters than the dimensions of the background space they occupy. 
As demonstrated in \cref{sample}, it is theoretically impossible to impose strict orthogonality on such layers.

To address this issue, we propose an alternative approach that enforces strict orthogonality on a subset of filters. 
Our method involves constructing an orthogonal structure within the background dimensions, and then allowing the remaining filters to be optimized without the constraint of orthogonality regularization. 
This approach helps prevent filter rank collapse in over-determined layers.
In the illustrated case, the filters reside in a 64-dimensional background space, where structural filters can span a maximum of 64 dimensions. 
We denote 64 structural filters in blue and the remaining relaxed 64 filters in red:

\begin{figure}[h]
  \centering
      \includegraphics[width= 0.85\textwidth]{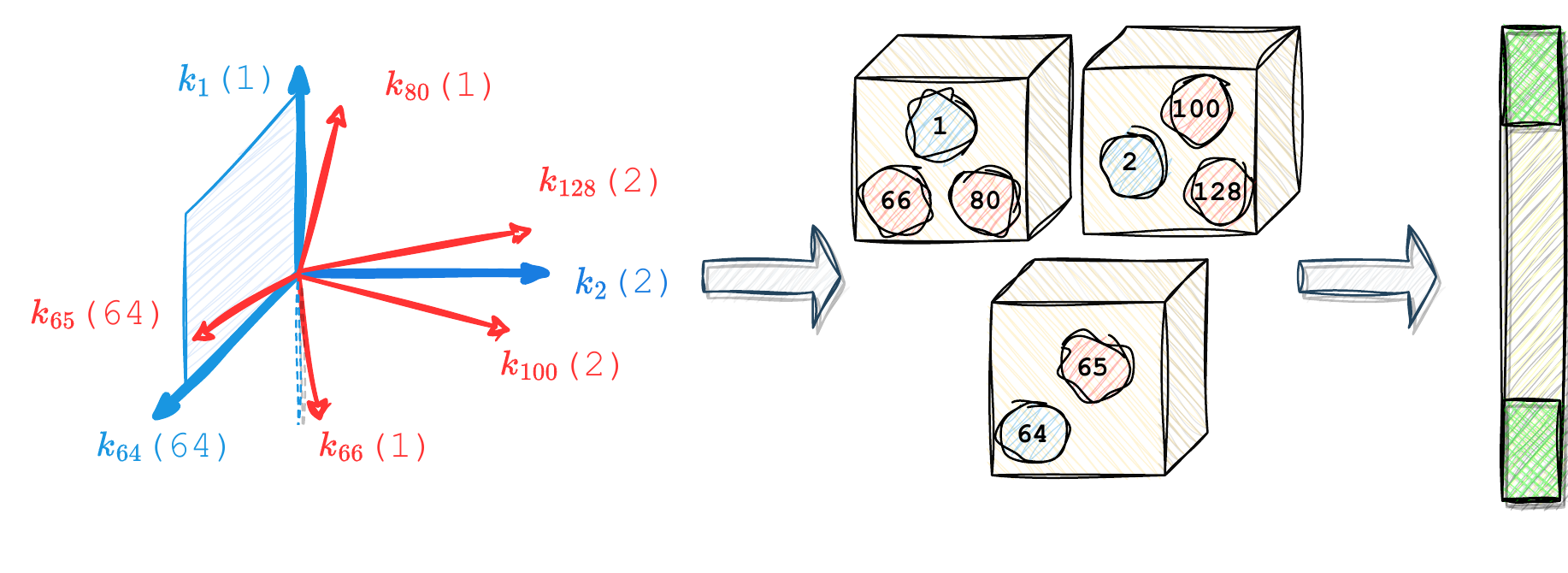}
      \caption{illustration for relaxed orthogonality regularization on overdetermined layers}
      \label{sample}
\end{figure}

For computational convenience, we consider applying relaxation to the previously generated correlation matrix. 
However, directly approximating orthogonality pairs in the correlation matrix is challenging. 
To illustrate this difficulty, we provide an example in the blue linear span in \cref{sample}. 
In this example, filter $k_{65}$ is not a structural filter that should be orthogonal to other structural filters, but it still lies within the blue $\operatorname{linear} \operatorname{span} \{k_{1}, k_{64}\} $ that is perpendicular to $k_2$.
Recognizing this challenge, we shift our focus and propose a rough approximation to determine the number of filter pairs that should be exempt from strict orthogonality regularization:

We assign labels to the freed filters based on the nearest structural filter, where the nearest is defined by minimal absolute correlation. 
We then assume that filters with the same label can have a high correlation with one another, 
and such high-correlation groups sharing the same label should be removed from strict orthogonality.
As the approximation number varies depending on the distribution of structural and freed filters, 
we employ Monte Carlo simulations to estimate the expected numbers for the relaxed high-correlation pairs.

In this case, we have 64 boxes representing the 64 structural filters. 
We then randomly assign the remaining 64 filters to these boxes, allowing us to count the number of high-correlation pairs (i.e., pairs within the same box). 
By repeating the random experiments, we approximate the expected number of relaxed correlation pairs within the lower triangle of the correlation matrix. 
After flattening the correlation lower triangular matrix, we sort it and remove the largest positive and negative correlation pairs, which share the relaxation number equally. 
This is represented by the green-colored section in the flattened correlation lower triangular matrix in \cref{sample}

\subsubsection{Relaxation on less-determined layers}

In this section, we propose the relaxation on less-determined layers. 
In our experiments, even when replacing the overdetermined layer with relaxed orthogonality regularization, 
the performance gain from strict orthogonality in deeper networks remains somewhat unsatisfactory. 
We guess the existence of a transition dimension that lies between the intrinsic dimension of data representation and the background dimension, determined by the network layers. 
This transition dimension, while higher than the intrinsic dimension of the dataset, increases slightly with model capacity. 
However, this increase does not perfectly align with the increase of background dimension as defined by the convolutional layer:

\begin{figure}[h]
  \centering
      \includegraphics[width= 0.85\textwidth]{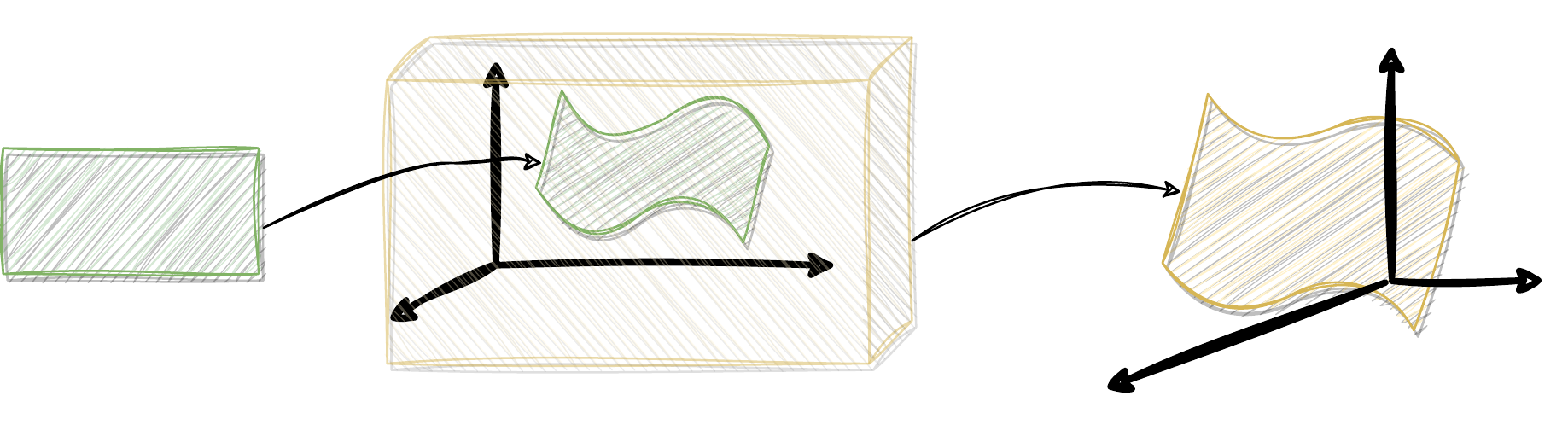}
      \caption{To illustrate the concept of the transition dimension, as we move from left to right: 
      The intrinsic dimension of data representation, represented in green, 
      the transition dimension, depicted in yellow. 
      Further, this transition dimension is embedded within the background dimension}
      \label{intrisinc_proper_background}
\end{figure}

As shown in the \cref{intrisinc_proper_background}, the intrinsic dimension of the dataset, can be embedded in a transition dimension, which may be lower than the background dimension. 
The data's intrinsic manifold can be represented effectively within this transition dimension. 
As model capacity increases, this transition dimension may increase slightly, but not to the extent that it matches the background dimension. 
Take WideResNet $28 \times 10$ and ResNet20 as an example, it is reasonable to expect that the transition dimension supported by WideResNet $28 \times 10$ would be larger than the proper span of ResNet20, but not ten times larger as the widenfactor of WideResNet.

The concept of a transition dimension, serving as a subset of the background dimension of the convolutional filters, 
introduces a fresh lens for understanding and applying orthogonality regularization. 
Instead of imposing strict orthogonality regularization on the entire background dimension, we focus on the transition dimension. 
This approach is referred to as relaxed disentangled orthogonality regularization on less-determined layers.

The question then arises: how to estimate the transition dimension of data representation? We adopt the following concepts to guide this estimation:

\begin{itemize}[leftmargin=10pt, itemsep=4pt]
  \item The intrinsic dimension, serves as a lower bound for the transition dimension.
  \item If a filter in the transition dimension directly corresponds to an attribute of the dataset, 
  the output of the convolutional filter layer could potentially serve as a "one-hot" vector. 
  The features highlighted in the output indicate the representation attribute.
  \item The transition dimension should vary depending on the model. For instance, the transition dimension of ResNet18 should be lower than that of ResNet50.
\end{itemize}

The approximation of the transition dimension is computed as follows:

\begin{equation}
\label{transition estimation}
  d_{\text{transition}} \leftarrow \min \left[ \max \left( \text{Attribute}, \text{Intrinsic} \right), \text{Max Transition} \right]
\end{equation}

The first part of the approximation is informed by the dataset, specifically the information it provides about the transition dimension of the incoming data representation. 
The second component, Max Transition, is determined by the model itself. 
As discussed earlier, the transition dimension is influenced by the model and will not increase proportionally with an increase of the background dimension of convolutional filters.

Furthermore, we propose an alternative for to alleviate the rank collapse in transition dimension. 
By optimized in a higher transition dimension, even if the rank collapse still happen, the problem is alleviated. 
In the convolutional module squences share the same background dimension in one layer,
we suggest progressively relaxing the transition dimension from the earlier to the later modules. 
This can be achieved by introducing a control hyperparameter to regulate the relaxation ratio within the correlation matrix. 
This modification provides an additional degree of flexibility in managing the challenges previously mentioned.
Detailed descriptions and demonstrations of this technique will be further elaborated in the experiment section \cref{hyperparameter scheme}.


\section{Experiments}
\label{experiments}


Our experimentation were conducted on the CIFAR100 and CIFAR10 datasets \cite{cifar_data}, both containing 60,000 $32 \times 32$ images. 
CIFAR100 and CIFAR10 are composed of 50,000 training images and 10,000 validation images, each containing 100 and 10 distinct labels, respectively. 
Adhering to the data splitting method proposed in \cite{resnet}, we divided the 50,000-image training set into a subset of 45,000 images for training and 5,000 for validation. 
The remaining 10,000 validation images as the test set.In our data preprocessing pipeline,
we applied a random crop transformation with a padding of 4 pixels to the 32x32 input images, 
followed by a random horizontal flip for data augmentation. 
These image tensors were normalized using predefined mean and standard deviation values.

Our experimentation involved a range of classical ResNet models \cite{resnet}, such as the narrow channel variants ResNet20, ResNet32, ResNet56, 
and the broader channel variants like ResNet18, ResNet34, as well as the ResNet50 model with a bottleneck structure. 
We also utilized the WideResNet 28$\times$10.

For training optimizer configuration, we followed the approach outlined in \cite{resnet}. 
On CIFAR10, we employed the SGD optimizer with a Nesterov Momentum of 0.9 for training the model over 160 epochs. 
The initial learning rate was set at 0.1 and reduced by a factor of 10 after the 80th and 120th epochs using MultiStepLR. 
For CIFAR100, we also used the SGD optimizer with a Nesterov Momentum of 0.9 to train the model over 200 epochs. 
The learning rate was initially set at 0.1, and then decreased by a factor of 5 after the 80th, 120th, and 160th epochs using MultiStepLR.
We set 128 for batchsize in SGD optimizer. In our experiments, we use 2 4090 RTX GPUs to train the model.


\subsection{Hyperparameter scheme}
\label{hyperparameter scheme}

In this section, we will introduce our scheme on the hyperparameter:

\begin{itemize}[leftmargin=10pt, itemsep=4pt]
  \item the hyperparameter on the balance of task loss and the loss of regularization terms
  \item the hyperparameter on the balance of diagonal loss and correlation loss 
  \item the hyperparameter on the relaxation of transition dimension
\end{itemize}

In determining the balance between task-specific loss (e.g., classification loss) and orthogonality regularization, 
we believe the key factor is the proportion each type of loss contributes to the total loss.
It is essential to ensure that the ratio of task-specific loss in the total loss does not become too small.
To determine the initial hyperparameter of orthogonality regularization,
We then adjust the hyperparameters of the regularization terms until this control statement is satisfied at Epoch 10.

\begin{equation}
  | \frac{\sum \text{Loss Regularization}}{\text{Total Loss}} - \text{Balance Regularization} | \leq \epsilon_\text{regularization}
\end{equation}

Here, we set $\text{Balance Regularization}$ to 10 $\%$ and $\epsilon_\text{regularization}$ to $1 \%$. 
During subsequent training, in reference to the Scheme Change for Regularization Coefficients \cite{srip}, 
we adjust our method at certain epochs. These include the beginning and midpoint of the second and third learning stages, as well as the start of the fourth learning stage. 
If the sum of the ratios of the regularization terms exceeds the set ratio, it is scaled down to less than $40 \%$. 
This strategy ensures a balanced and flexible approach to model training, adjusting the contribution of different loss components as training progresses.

When balancing diagonal loss and correlation loss, we prioritize correlation loss as it plays a more critical role in orthogonality regularization. 
Similar to the task balance control, at the same epoch where we monitor the balance between task-specific loss and regularization loss, 
we set the balance between diagonal loss and correlation loss to be $10 \%$ and $\epsilon_\text{disentangled}$ to $ 5 \%$.



In terms of the hyperparameters for relaxing the transition dimension, we adopt a progressive relaxation approach from the earlier to the later modules. 
By employ a control ratio datamap in the range of $[0,1]$ to govern the number of green (relaxed) filter pairs in the correlation matrix, as shown in \cref{sample}.
As the control ratio increases, the transition dimension decreases accordingly.

All convolutional filters are divided into groups according to their filter number and background space dimensions. 
To allow for a larger transition dimension, layers that appear earlier within the same group are assigned a smaller control ratio. 
This methodology ensures a balanced transition dimension across the network while mitigating potential rank collapse issues in the later layers.

\subsection{On the performance gains under orthogonality regularization}

In the following sections, we examine the impact of orthogonality regularization. 
Relaxed disentangled orthogonality techniques are applied by default to the overdetermined layers in both strict and relaxed disetangled orthogonality.
For the addtional hyperparameters introduced by relaxed orthogonality in different models:

\begin{itemize}[leftmargin=10pt, itemsep=4pt]
  \item For the intrinsic dimension dimension, we refer the research of \cite{intrinsic_dimension}, set intrinsic dimension as 30 in our experiments. The attribute of CIFAR100 is 100, 10 for CIFAR10.
  \item For the narrow-width ResNet, such as ResNet20, the Max Transition was set to 30 with no control ratio map applied. 
  \item The mid-width ResNet, like ResNet18, had a Max Transition of 60, while the network with bottleneck structure had a Max Transition of 80.
  \item For WideResNet models, the Max Transition was set to 100.
\end{itemize}

Upon analyzing narrow ResNet models, we found that strict disentangled orthogonality could impede performance in shallow variants, 
which are the models with the lowest background dimension in our experiment. 
However, the introduction of the transition dimension in narrow ResNet models can still enhance their performance. 
As the network depth increased, the advantage of strict orthogonality regularization on the background dimension became evident. 
Moreover, relaxed orthogonality regularization on the transition dimension led to further performance improvement.

In the case of mid-width ResNet models, an increased background dimension, induced by a higher number of filters, creates a more complex background dimension. 
Baseline regularization methods like Frobenius and SRIP improved model performance, and the application of relaxed orthogonality showed its advantage in shallow models like ResNet. 
In comparison, the overdetermined layer in the bottleneck structure seemed to challenge the effectiveness of strict orthogonality regularization methods. 
However, our proposed relaxation on overdetermined layers stabilized the training process and led to superior performance in ResNet models with bottleneck structures.

For WideResNet models, both strict orthogonality regularization on the background dimension and relaxed orthogonality on the transition dimension demonstrated their benefits. 
However, strict disentangled orthogonality regularization on the background dimension appeared to be the least effective for performance improvement.
This might be due to the fact that WideResNet models have the highest background dimension in our experiment, 
making the introduction of the transition dimension in WideResNet models very significant.

\begin{table*}[htbp]
  \caption{\label{outcome-table} 
  The table presents the test accuracy outcomes for different cases, displayed as mean and standard deviation values derived from three runs with random seeds. 
  Different orthogonality regularization methods are listed along the rows. 
  The term Vanilla refers to optimization without regularization, 
  Strict indicates strict disentangled orthogonality in the background space, 
  and Relaxed represents relaxed disentangled orthogonality in the transition dimension estimated by \cref{transition estimation}. 
  WRN 28$\times$10 in the last row represents WideResNet 28$\times$10.}
  \scalebox{0.88}{
  \parbox{.5\linewidth}{
  \setlength{\tabcolsep}{3.75mm}{
      \begin{tabular}{@{}cccccc@{}}
      \hline
      \textbf{\textit{Test Acc Mean/Std}}                                     &  \textbf{\textit{Vanilla}}                                 & \textbf{\textit{Frobenius}}                        & \textbf{\textit{SRIP}}                                & \textbf{\textit{Strict}}       & \textbf{\textit{Relaxed}}  \\ \hline
      \multicolumn{6}{c}{\textit{16-32-64}} \\
      \hline 
      \textbf{\textit{ResNet20}}                                     & 91.65 $\pm$ 0.15                            & 91.68 $\pm$0.11                 & 91.75$\pm$0.15                   & 91.57$\pm$0.12                            & \textbf{91.88$\pm$0.11}                              \\ 
      \textbf{\textit{ResNet32}}                                     & 92.81 $\pm$ 0.21                            & 92.81$\pm$0.12                  & 92.85$\pm$0.14                   & 92.71$\pm$0.18                            & \textbf{93.02$\pm$0.17}                             \\ 
      \textbf{\textit{ResNet56}}                                     & 93.25 $\pm$ 0.17                            & 93.30$\pm$0.16                  & 93.47$\pm$0.15                   & 93.15$\pm$0.19                             & \textbf{93.51$\pm$0.08}                                \\ \hline
      \multicolumn{6}{c}{\textit{64-128-256-512}} \\
      \hline
      \textbf{\textit{ResNet18}}                                     & 76.51$\pm$0.18                                 & 76.87$\pm$0.13                    & 77.10$\pm$0.18                    & 77.07$\pm$ 0.16                   & \textbf{77.33$\pm$0.10}                                      \\ 
      \textbf{\textit{ResNet34}}                                     & 77.08$\pm$0.22                                 & 77.43$\pm$0.16                    & 77.69$\pm$0.12                    & 77.61$\pm$ 0.18                   & \textbf{77.83$\pm$0.16}                                      \\ 
      \textbf{\textit{ResNet50}}                                     & 77.43$\pm$0.16                                 & 77.82$\pm$0.17                    & 77.71$\pm$0.22                    & 78.10$\pm$ 0.12                   & \textbf{78.48$\pm$0.15}                                      \\ \hline
     \multicolumn{6}{c}{\textit{160-320-640}} \\
      \hline
      \textbf{\textit{WRN 28$\times$10}}                            & 79.32 $\pm$ 0.16                      & 79.82 $\pm$ 0.13           & 80.11 $\pm$ 0.12                  & 79.71 $\pm$ 0.21                        & \textbf{80.21$\pm$0.11}                              \\ \hline
  \end{tabular}}}}
\end{table*}

\subsection{On the near-orthogonality under orthogonality regularization}
\label{near-orth-section}

In this section, we will examine the extent of near-orthogonality under various orthogonality regularizations. 
We will focus on the following models: narrow variants of ResNet (ResNet56), mid-width variants of ResNet (ResNet18), and the wider variant WideResNet (WRN28$\times$10).
For the less-determined layers, we exhibite the average statistics of all layers in the same background dimension.

\begin{table*}[htbp]
  \caption{\label{near-orth-table} In the table, we quantify the near-orthogonality of a specific layer by analyzing the statistics of the correlation matrix and the diagonal. 
  The mean of the lower triangular part of the correlation matrix represents the average degree to which filters in a specific transformation approach zero-correlation. 
  And the following the standard deviation of correlation indicates the stability of near-orthogonality. 
  Separated by '/', recording the average diagonal in the layer}
  \scalebox{0.88}{\parbox{.5\linewidth}{%
      \begin{tabular}{@{}cccccc@{}}
      \hline
      \textbf{\textit{ResNet56}}                              &  \textit{Layer3 Downsample}                                & \textit{ Layer1 [16,144]}                          & \textit{ Layer2 [32,288]}                           & \textit{Layer3 [64,576]}                                   \\ \hline
       \textbf{\textit{Vanilla}}                                     &  0.01 $\pm$ 0.10/0.03                          & 0.04 $\pm$ 0.25/0.06                  &  0.01 $\pm$  0.11/0.05                  &  0.01 $\pm$  0.06/0.18                                    \\ 
       \textbf{\textit{Frobenius}}                                   & 0.02 $\pm$ 0.19/0.13                           & -0.00 $\pm$ 0.05/0.27                   & -0.00 $\pm$  0.09/0.22                    & -0.01 $\pm$  0.09/0.42                               \\ 
       \textbf{\textit{SRIP}}                                        & 0.01 $\pm$ 0.18/0.17                          & 0.00 $\pm$ 0.02/0.23                    & 0.00 $\pm$  0.09/0.24                     & -0.01 $\pm$  0.09/0.45                                \\ 
       \textbf{\textit{Strict}}                                      & 0.01 $\pm$ 0.13/0.17                  & \textbf{0.00 $\pm$ 0.00/0.95}           & \textbf{-0.00 $\pm$  0.01/0.90}           & \textbf{0.00 $\pm$  0.01/1.12}                                   \\ 
       \textbf{\textit{Relaxed}}                                     & \textbf{0.00 $\pm$ 0.13/0.20}                           & 0.00 $\pm$ 0.01/0.31                    & -0.00 $\pm$  0.02/0.30                   & -0.00 $\pm$  0.03/0.60                                  \\ \hline
       \textbf{\textit{ResNet18}}                              &  \textit{Layer3 Downsample}                   &  \textit{Layer2 [128,1152]}                       &  \textit{Layer3 [256,2304]}                          &  \textit{Layer4 [512,4608]}                                 \\ \hline
     \textbf{\textit{Vanilla}}                                     &  0.01 $\pm$ 0.10/0.03                          & 0.04 $\pm$ 0.25/0.06                  &  0.01 $\pm$  0.11/0.05                  &  0.01 $\pm$  0.06/0.16                                    \\ 
       \textbf{\textit{ Strict}}                                      & \textbf{0.01 $\pm$ 0.10/0.11}                  & \textbf{0.00 $\pm$ 0.01/0.26}         & \textbf{0.00 $\pm$  0.02/0.16}          &  \textbf{0.01 $\pm$  0.02/0.18}                              \\ \hline
       \textbf{\textit{WRN 28 × 10}}                          &  \textit{Layer3 Downsample}                 &  \textit{Layer1 [160,1440]}                        &  \textit{Layer2 [320,2880]}                          &  \textit{Layer3 [640,5760]}                                           \\ \hline
     \textbf{\textit{ Vanilla}}                                     &  0.01 $\pm$ 0.10/0.03                          & 0.04 $\pm$ 0.25/0.06                  &  0.01 $\pm$  0.11/0.05                  &  0.01 $\pm$  0.06/0.18                                    \\ 
       \textbf{\textit{Strict}}                                     & \textbf{0.01 $\pm$ 0.05/0.03}                  & \textbf{0.00 $\pm$ 0.05/0.06}         &  \textbf{0.00 $\pm$  0.03/0.06}         &  \textbf{0.01 $\pm$  0.05/0.31}                             \\ \hline
      \end{tabular}
  }}
\end{table*} 

Notably, no regularization scheme can achieve perfect orthogonality in the less-determined layers of the well-trained models. 
Starting with the narrowest ResNet variant, ResNet56, due to its low-dimension background space, the well-trained model under strict disentangled orthogonality almost achieves perfect orthogonality. 
However, this near-orthogonality property diminishes significantly with the increase in the dimension of the background dimension.  
The higher the dimension of the background dimension, the less likely it is that the manifold associated with a good task loss overlaps with the manifold exhibiting near-orthogonality.

\section{Summary}

\subsection{Rethinking strict orthogonality regularization}

Given our observations on near-orthogonality \cref{near-orth-table} and the improvement in model performance \cref{outcome-table}, it becomes clear that strict orthogonality regularization may not be the optimal approach. 
Even though ResNet56 can achieve both near-orthogonality in the well-trained model and a performance gain from strict orthogonality regularization on the background dimension, 
it still falls short when compared to relaxed orthogonality regularization on the transition dimension. 
Therefore, we should reconsider the prevailing principle of adhering to strict orthogonality regularization on the background dimension.

\subsection{Limitations}

Our approach has certain limitations that should be addressed.

\begin{itemize}[leftmargin=10pt, itemsep=4pt]
\item the estimation of the transition dimension could benefit from a more theoretically grounded method. 
The need for a control ratio in the earlier layers might be a consequence of an inaccurate approximation of the true dimension of the transition space.
Ideally, we could develop models that allow for module-wise relaxation configurations, thereby providing greater flexibility and control over the training process. 
However, implementing such a feature in practice may pose its own challenges.
\item  There's the issue of computational complexity. 
The introduction of a double search for the positive and negative boundaries of the relaxation correlation filter pair means that the relaxed orthogonality regularization tends to be more computationally intensive than other regularization methods. 

\end{itemize}

\newpage

\bibliographystyle{plainnat}
\bibliography{neurips_2023}

\newpage
\section{Supplementary Material}

The first subsection is dedicated to discussing the motivation behind the introduction of the disentangled norm,
The second subsection explores the potential negative impact of enforcing strict orthogonality during the training process of deeper models
In the third subsection, we delve into the consequences of extreme relaxation on the transition dimension, along with a discussion of the relaxed orthogonality regularization training.
In the final subsection, we conduct an exploratory experiment, assessing whether orthogonality regularization creates a trade-off between model robustness and precision through an out-of-distribution test.

\subsection{Why traditional measure disordered in near orthogonality}

In the discussion by \citet{srip}, the issue of over-determined input is revealed as an inaccessible problem setting for conventional less-determined orthogonality regularization. 
Moreover, we wish to elaborate on another two specific issues in traditional measures, using the Frobenius norm as an example:

\begin{itemize}[leftmargin=10pt, itemsep=4pt]
\item The imbalance between the residuals of diagonal norm and the residuals of lower triangular triangle
\item The unfair evaluation of correlation magnitude in gram-based orthogonality measure
\end{itemize}

The Gram matrix can be dissected into two parts, highlighting the imbalance:

\begin{align}
  K K^{\top}
  = 
  \left[\begin{array}{cccc}
  0 & k_1^{\top}k_2 & \cdots & k_1^{\top}k_n \\
  k_2^{\top}k_1 & 0 & & k_2^{\top}k_n \\
  \vdots & & \ddots & \vdots \\
  k_n^{\top}k_1 & k_n^{\top}k_2 & \cdots & 0
  \end{array}\right]
  + 
  \left[\begin{array}{ccc}
  k_1^{\top}k_1  & \cdots & 0 \\
  \vdots         & \ddots & \vdots \\
  0              & \cdots & k_n^{\top}k_n
  \end{array}\right]
\end{align}

Conventional orthogonal regularization, such as the Frobenius norm, 
aims to converge the $n$ diagonal elements of the Gram matrix (termed \emph{diag}) to $1$ while pulling the $\frac{n(n-1)}{2}$ lower triangular triangle elements (termed \emph{tril}) to $0$. 
This approach effectively merges two fundamentally different optimization targets within a single Frobenius norm. 
To illustrate, we can reconstruct the Frobenius norm as:

\begin{equation}
  \left\|K K^{\top}-I\right\|_F=\sqrt{\sum_i^n\left(\emph{diag}_i-1\right)^2+2 \sum_j^{\frac{n(n-1)}{2}} \emph{tril}_j ^2}
\end{equation}

In typical settings where the identity is accessible for the Gram matrix, 
the number imbalance between \emph{diag} and \emph{tril} elements doesn't cause a problem. 
However, as demonstrated in previous near-orthogonality table, 
in the shortcut representing over-determined cases, \emph{tril} part will dominate the Frobenius norm. 
Conversely, in less-determined cases, \emph{diag} will contributs more to the loss. 
Mixing two distinct loss patterns to be optimized within single Frobenius norm seems inappropriate.

Beyond the imbalance number issue, 
we have argued that the zero correlation matrix should be the central goal in orthogonality regularization. 
As discussed earlier, in the strictly orthogonality-constrained Gram matrix $K K^{\top}$, 
it should coincide with the identity matrix $I$, indicating zero correlation between the filters and equal norms for all filters. 
However, this ideal is not well-represented by the gram-based orthogonality regularization. 
Discrepancies between the diagonal norm and the identity lead to an unfair evaluation of correlation information, resulting in two potential problems:

\begin{itemize}[leftmargin=10pt, itemsep=4pt]
  \item Under the condition $\|k_1\| \neq \|k_2\|$, if $k_3^{\top} k_1 = k_3^{\top} k_2$, the correlations between these two elements could differ substantially
  \item For $k_1^{\top} k_2$, under the assumption that $\|k_1\|, \|k_2\| \leq 1$, the model might undervalue the magnitude of $k_1^{\top} k_2$ and halt optimization prematurely
\end{itemize}
  
In typical optimization scenarios, the aforementioned problems may not manifest. However, the presence and dominance of the loss task, 
restrict the search space for the orthogonality regularization term. This limitation is clearly demonstrated in inaccessible orthogonality illustration of the main context.

To address these issues, we adopt the following strategies:

\begin{itemize}[leftmargin=10pt, itemsep=4pt]
\item We bifurcate the input convolutional into two categories: the over-determined and the less-determined cases. For the over-determined layers, we enforce our proposed relaxed orthogonality.
\item Instead of using gram-based orthogonality regularization, which is influenced by the magnitude of filter norms, we introduce a correlation-based approach. This new method solely focuses on the correlation between the filters, thus, making it independent of their filter magnitudes.
\end{itemize}

\subsection{Implications of strict orthogonality on the training}

In this section, we unravel the conceptual conflict stemming from opposing gradients introduced by task loss and strict orthogonality within the background space:

Consider the background space, spanned by the black coordinate filters. 
Within this space, the green-shadowed transition dimension is embedded, which is effectively spanned by a set of linear filters $\{k_1, k_2, \ldots, k_n \}$. 
However, due to the higher dimensionality of the background space compared to the transition dimension, we observe "redundant" transition dimension filters, such as $k_{n+1}$.

\begin{figure}[!htbp]
  \centering
  \includegraphics[width= 0.5\textwidth]{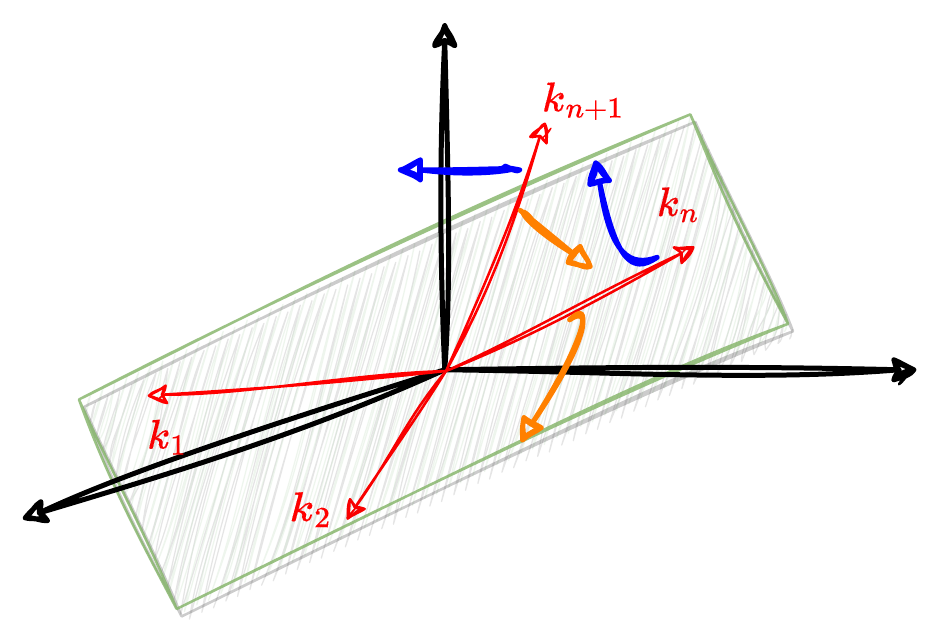}
  \caption{Depiction of the conflict between strict orthogonality and task loss}
  \label{rank_curse}
\end{figure}

During optimization, a conflict emerges involving the structural filters for the transition dimension and task filters. 
Let's delve deeper into this conflict by examining filters $k_n$ and $k_{n+1}$:

\begin{itemize}[leftmargin=10pt, itemsep=4pt]
  \item From the perspective of strict orthogonality regularization, blue gradients are imposed that strive to span a larger transition dimension. 
  This results in the extraction of in-span filter $k_n$ from the transition dimension and the orthogonalization of $k_{n+1}$, 
  regardless of the existing linear span of the transition dimension
  \item On the other hand, the task loss introduces orange gradients on $k_n$ and $k_{n+1}$, possibly drawing filters into the current transition dimension. 
  Simultaneously, it rearranges the filter distribution within the transition dimension to optimize the layer-wise data representation.
\end{itemize}

When focusing solely on strict orthogonality in the background dimension, issues arise if the blue gradients become too strong, leading to over-regularization of orthogonality. 
This may inadvertently result in a wastage of filters, given that the existing linear span of the transition dimension can effectively represent most of the input.

Our proposed resolution involves relaxing some highly correlated pairs like $( k_n, k_{n+1} )$ from the correlation orthogonality regularization. 
We hypothesize that such relaxation can assuage the conflicts in the imagined transition dimension.

\subsection{The module-wise relaxation on the transition dimension}

In the previous section, we show how strict orthogonality hinder the training of deeper layers. 
While excessive relaxation in the transition dimension can present its own set of challenges, such as rank collapse in the imagined transition dimension. 
For example, if we enforce strict orthogonality only on a 0-dimension transition dimension, 
this actually would be equivalent to not imposing any strict orthogonality regularization on the transition dimension, 
thereby highlighting the detrimental effects of a underestimated transition dimension.

To guard against the rank collapse that can stem from inaccurate approximation of the transition dimension, 
we propose a gradual reduction method for the transition dimension. 
Notably, our proposal primarily affects the main pathway of the network, excluding the overdetermined layers.

\begin{figure}[!htbp]
  \centering
  \includegraphics[width= 0.85\textwidth]{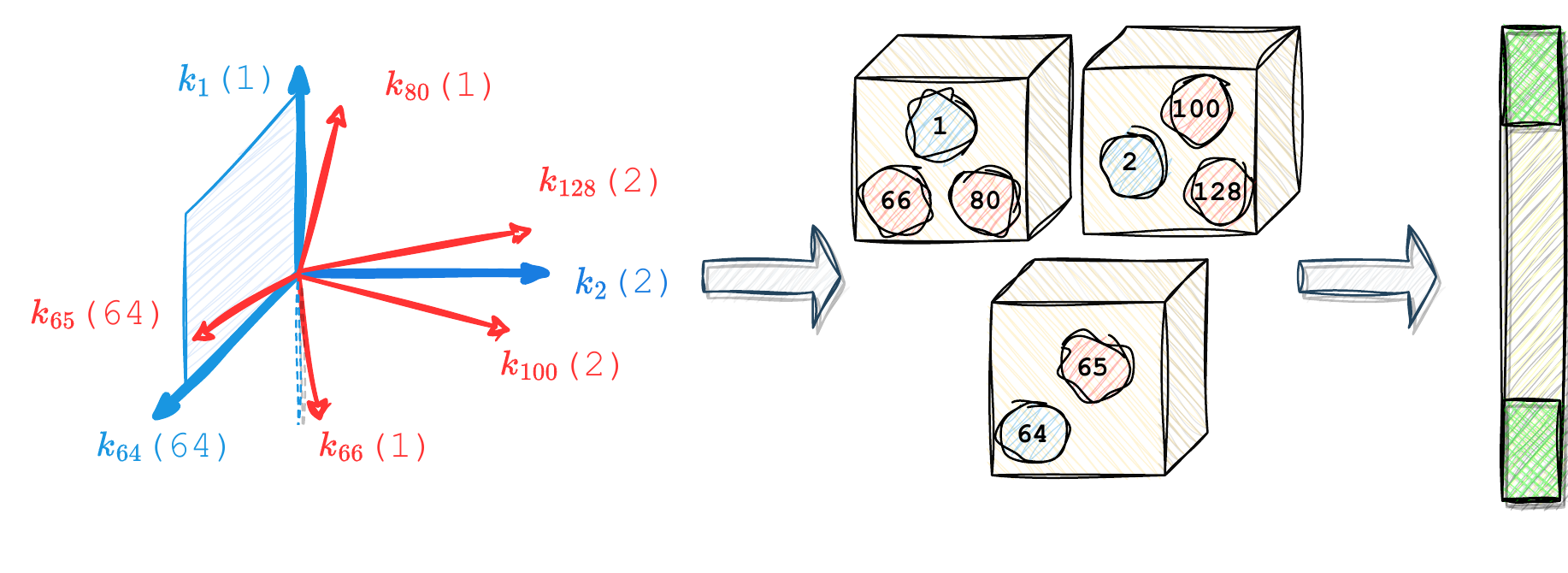}
  \caption{In contrast to the main context, which refers to the relaxation in the overdetermined layers, this figure presents the relaxation in the less-determined layers.}
  \label{sample_less_determined}
\end{figure}



In the less-determined layers, a ratio map, ranging from $0$ to $1$, is introduced for the green relaxation filters as illustrated in Figure \ref{sample_less_determined}.
This map signifies how many high-correlation estimated relaxation pairs are exempted from strict orthogonality. 
A value of $1$ indicates that all such pairs are exempted, while a value of $0$ means that despite the existence of some estimated relaxation pairs, none are excluded from strict orthogonality during the training process.

\begin{figure}[!htbp]
  \centering
  \includegraphics[width= 0.8\textwidth]{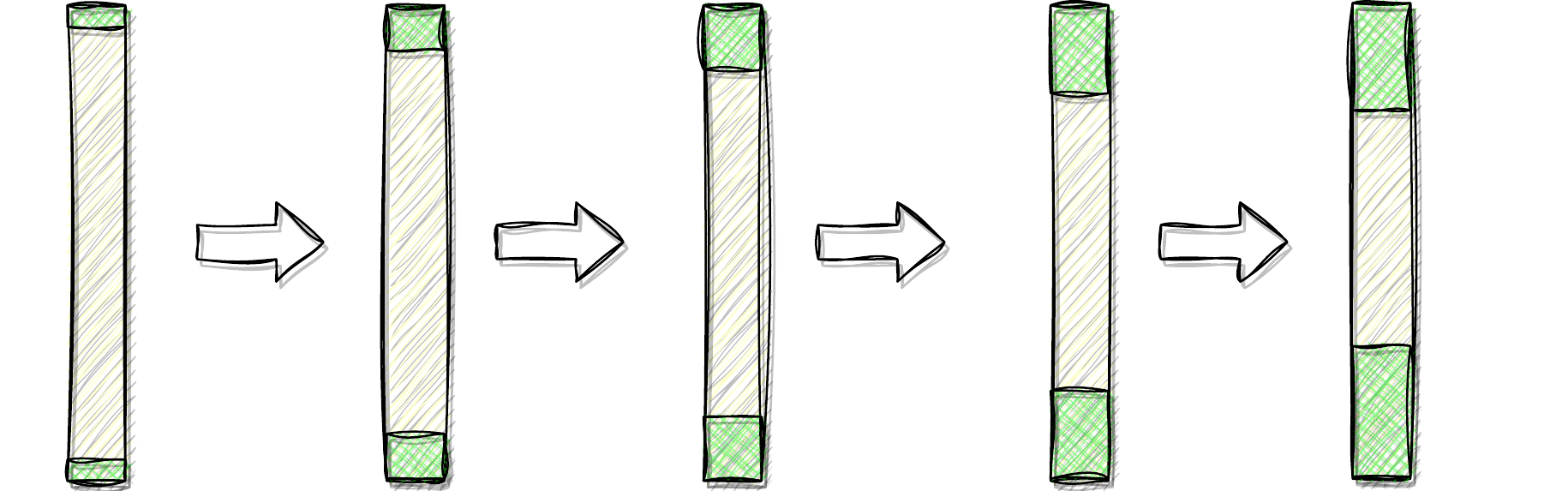}
  \caption{Applying layer-wise ratio map in the convolutional layers to prevent rank collapse, particularly in the transition dimension of the former layer}
  \label{module_wise}
\end{figure}

If rank collapse in the transition occurs in the earlier convolution modules, subsequent layers face a dual challenge. 
They must deal with the rank collapse within their filters, as well as the data representation emanating from earlier layers already subjected to rank collapse. 
As demonstrated in \cref{module_wise}, 
earlier convolutional modules in the less-determined layer, having identical filter numbers, will exclude fewer ratio pairs from strict orthogonality in order to mitigate rank collapse in the transition dimension.

We introduce a hyperparameter and three interploation patterns for the layerwise ratio map:

\begin{itemize}[leftmargin=10pt, itemsep=4pt]
  \item \emph{least\_ratio}: the least relaxation ratio for the first layer (the formest layer)
  \item \emph{linear}: linear interploation from \emph{least\_ratio} to $1$ in the finalist layer:
    \begin{equation}
      \emph{ratio\_map}_i = \emph{least\_ratio} + (1 - \emph{least\_ratio})*\frac{\emph{i}}{\emph{module \_ number}}
    \end{equation}
  \item \emph{log}: logarithm interploation from \emph{least\_ratio} to $1$ in the finalist layer 
  \begin{equation}
    \emph{ratio\_map}_i = \emph{least\_ratio} + (1 - \emph{least\_ratio}) * \frac{\log(1 + \emph{i})}{\log(1 + \emph{module \_ number})}
  \end{equation}
  \item \emph{exp}: expenotial interploation from \emph{least\_ratio} to $1$ in the finalist layer 
  \begin{equation}
    \emph{ratio\_map}_i = \emph{least \_ ratio} + (1 - \emph{least \_ ratio}) * \left(1 - \exp\left(-\frac{\emph{i}}{\emph{module \_ number}}\right)\right)
  \end{equation}
\end{itemize}

Given the same \emph{least \_ ratio}, the logarithm interpolation pattern provides the tightest constraint on the transition dimension of subsequent layers. 
By default, we use logarithm interpolation in our experiments.

We present examples contrasting the unmodified module-wise relaxation ratio with the modified relaxation map. Table \ref{ratio map comparsion} 
compares the module-wise relaxation ratio data map for WideResNet $28 \times 10$.

\begin{table}
  \centering
  \caption{The comparison for module-wise relaxation ratio datamap for WideResNet $28 \times 10$, left one is the }
  \begin{minipage}{0.3\textwidth}
      \centering
      \begin{tabular}{|l|r|}
          \hline
          \textbf{Layer} & \textbf{Value} \\
          \hline
          stem.conv1 & 0.0249 \\ \hline
          layer1.0.conv1 & 0.0028 \\
          layer1.0.conv2 & 0.0028 \\
          layer1.0.downsample.0 & 0.0247 \\
          layer1.1.conv1 & 0.0028 \\
          layer1.1.conv2 & 0.0028 \\
          layer1.2.conv1 & 0.0028 \\
          layer1.2.conv2 & 0.0028 \\
          layer1.3.conv1 & 0.0028 \\
          layer1.3.conv2 & 0.0028 \\ \hline
          layer2.0.conv1 & 0.0040 \\
          layer2.0.conv2 & 0.0040 \\
          layer2.0.downsample.0 & 0.0036 \\
          layer2.1.conv1 & 0.0040 \\
          layer2.1.conv2 & 0.0040 \\
          layer2.2.conv1 & 0.0040 \\
          layer2.2.conv2 & 0.0040 \\
          layer2.3.conv1 & 0.0040 \\
          layer2.3.conv2 & 0.0040 \\ \hline
          layer3.0.conv1 & 0.0050 \\
          layer3.0.conv2 & 0.0050 \\
          layer3.0.downsample.0 & 0.0039 \\
          layer3.1.conv1 & 0.0050 \\
          layer3.1.conv2 & 0.0050 \\
          layer3.2.conv1 & 0.0050 \\
          layer3.2.conv2 & 0.0050 \\
          layer3.3.conv1 & 0.0050 \\
          layer3.3.conv2 & 0.0050 \\
          \hline
      \end{tabular}
  \end{minipage}
  \hspace{0.05\textwidth} 
  \begin{minipage}{0.3\textwidth}
      \centering
      \begin{tabular}{|l|r|}
          \hline
          \textbf{Layer} & \textbf{Value} \\
          \hline
          stem.conv1 & 0.0249 \\ \hline
          layer1.0.conv1 & 0.0009 \\
          layer1.0.conv2 & 0.0010 \\
          layer1.0.downsample.0 & 0.0247 \\
          layer1.1.conv1 & 0.0013 \\
          layer1.1.conv2 & 0.0016 \\
          layer1.2.conv1 & 0.0018 \\
          layer1.2.conv2 & 0.0020 \\
          layer1.3.conv1 & 0.0023 \\
          layer1.3.conv2 & 0.0027 \\ \hline
          layer2.0.conv1 & 0.0014 \\
          layer2.0.conv2 & 0.0015 \\
          layer2.0.downsample.0 & 0.0036 \\
          layer2.1.conv1 & 0.0020 \\
          layer2.1.conv2 & 0.0023 \\
          layer2.2.conv1 & 0.0026 \\
          layer2.2.conv2 & 0.0030 \\
          layer2.3.conv1 & 0.0034 \\
          layer2.3.conv2 & 0.0039 \\ \hline
          layer3.0.conv1 & 0.0015 \\
          layer3.0.conv2 & 0.0017 \\
          layer3.0.downsample.0 & 0.0039 \\
          layer3.1.conv1 & 0.0022 \\
          layer3.1.conv2 & 0.0026 \\
          layer3.2.conv1 & 0.0030 \\
          layer3.2.conv2 & 0.0035 \\
          layer3.3.conv1 & 0.0041 \\
          layer3.3.conv2 & 0.0049 \\
          \hline
      \end{tabular}
  \end{minipage}
  \label{ratio map comparsion}
  \end{table}


\subsection{Orthogonality Regularization: A trade-off between robustness and precision? An out-of-distribution test}

In this section, we discuss the results of experiments conducted on CIFAR10 and CIFAR100 datasets, 
wherein we employ an out-of-distribution test set created through a 15-degree rotation (using PyTorch's rotation functionality) on the original test set. 
For these tests, the filter extractors and classifiers of the models remain unmodified.

\begin{table*}[!htbp]
  \caption{\label{rotation_outcome-table} 
  The table presents the test accuracy outcomes for different cases, displayed as mean and standard deviation values derived from three runs with random seeds. 
  Different orthogonality regularization methods are listed along the rows. 
  The term Vanilla refers to optimization without regularization, 
  Strict indicates strict disentangled orthogonality in the background space, 
  and Relaxed represents relaxed disentangled orthogonality in the transition dimension.
  WRN 28$\times$10 in the last row represents WideResNet 28$\times$10.}
  \scalebox{0.88}{
  \parbox{.5\linewidth}{
  \setlength{\tabcolsep}{3.75mm}{
      \begin{tabular}{@{}cccccc@{}}
      \hline
      \textbf{\textit{Test Acc Mean/Std}}                                     &  \textbf{\textit{Vanilla}}                                 & \textbf{\textit{Frobenius}}                        & \textbf{\textit{SRIP}}                                & \textbf{\textit{Strict}}       & \textbf{\textit{Relaxed}}  \\ \hline
      \multicolumn{6}{c}{\textit{16-32-64}} \\
      \hline 
      \textbf{\textit{ResNet20}}                                     & \textbf{83.53 $\pm$ 0.23}                            & 82.79 $\pm$ 0.21                  & 83.01 $\pm$ 0.16                   & 82.29 $\pm$ 0.31                            & 83.39 $\pm$ 0.21                             \\ 
      \textbf{\textit{ResNet32}}                                     & 84.63 $\pm$ 0.28                              & 84.52 $\pm$ 0.32                  & \textbf{84.81 $\pm$ 0.24}                   & 84.21 $\pm$ 0.28                            & 84.66 $\pm$ 0.25                              \\ 
      \textbf{\textit{ResNet56}}                                     & 85.33 $\pm$ 0.19                              & 84.93 $\pm$ 0.22                  & 85.31 $\pm$ 0.21                   & 84.95 $\pm$ 0.25                            & \textbf{85.45 $\pm$ 0.14}                               \\ \hline
      \multicolumn{6}{c}{\textit{64-128-256-512}} \\
      \hline
      \textbf{\textit{ResNet18}}                                     & 64.22 $\pm$ 0.28                                 & 64.50 $\pm$ 0.25                    & 64.65 $\pm$ 0.22                    & 64.87 $\pm$ 0.19                   & \textbf{65.23 $\pm$ 0.13}                                      \\ 
      \textbf{\textit{ResNet34}}                                     & 65.21 $\pm$ 0.19                                 & 65.50 $\pm$ 0.21                    & 65.97 $\pm$ 0.19                    & 65.87 $\pm$ 0.21                   & \textbf{66.13 $\pm$ 0.18}                                        \\ 
      \textbf{\textit{ResNet50}}                                     & 65.82 $\pm$ 0.26                                 & 65.81 $\pm$ 0.19                    & 66.16 $\pm$ 0.15                    & 66.46 $\pm$ 0.17                   & \textbf{66.70 $\pm$ 0.19}                                      \\ \hline
     \multicolumn{6}{c}{\textit{160-320-640}} \\
      \hline
      \textbf{\textit{WRN 28$\times$10}}                            & 67.41 $\pm$ 0.13                                 & 67.71 $\pm$ 0.18                    & 67.56 $\pm$ 0.21                    & 67.66 $\pm$ 0.19                   & \textbf{67.93 $\pm$ 0.15}                          \\ \hline
  \end{tabular}}}}
\end{table*} 

Insights drawn from the narrow-filter ResNet models reveal that the introduction of orthogonality regularization adds complexity to the out-of-distribution test set, particularly with strict orthogonality methods. 
As the model depth increases, these narrow-width ResNet models begin to derive benefits from orthogonality regularization.
However, these benefits from orthogonality regularization remain relatively modest or insignificant. 
Therefore, the application of orthogonality regularization on narrow-width ResNet may need consideration.

For medium-width ResNet and WideResNet models, 
it becomes evident that orthogonality regularization not only improves the outcomes of the original test set, but also enhances the training outcomes of the out-of-distribution test set. 
This suggests that orthogonality regularization is not a trade-off between robustness and precision, but rather, it reduces the wastage of model filters that occurs under training without regularization. 
This phenomenon may be attributed to the tendency of non-regularized training to get stuck in local minima as model capacity (or background dimension) increases. 
Furthermore, by introducing relaxed orthogonality at an appropriate transition dimension, we can more effectively squeeze the model's capacity!

\end{document}